\documentclass[10pt,twocolumn,letterpaper]{article}

\usepackage{wacv}
\usepackage{times}
\usepackage{epsfig}
\usepackage{graphicx}
\usepackage{amsmath}
\usepackage{amssymb}
\usepackage{booktabs}

\usepackage[accsupp]{axessibility}  

\usepackage{xcolor}

\def\wacvPaperID{655} 

\wacvapplicationstrack 

\wacvfinalcopy 

\ifwacvfinal
\usepackage[breaklinks=true,bookmarks=false]{hyperref}
\else
\usepackage[pagebackref=true,breaklinks=true,colorlinks,bookmarks=false]{hyperref}
\fi

\begin{document}

\title{Self-Supervised Learning with Masked Image Modeling for Teeth Numbering, Detection of Dental Restorations, and Instance Segmentation in Dental Panoramic Radiographs}
 
\author{Amani Almalki \qquad Longin Jan Latecki\\
Department of Computer and Information Sciences, Temple University, Philadelphia, USA\\
{\tt\small \{amani.almalki,latecki\}@temple.edu}
}

\maketitle
\thispagestyle{empty}

\begin{abstract}
   The computer-assisted radiologic informative report is currently emerging in dental practice to facilitate dental care and reduce time consumption in manual panoramic radiographic interpretation. However, the amount of dental radiographs for training is very limited, particularly from the point of view of deep learning. This study aims to utilize recent self-supervised learning methods like SimMIM and UM-MAE to increase the model efficiency and understanding of the limited number of dental radiographs. We use the Swin Transformer for teeth numbering, detection of dental restorations, and instance segmentation tasks. To the best of our knowledge, this is the first study that applied self-supervised learning methods to Swin Transformer on dental panoramic radiographs. Our results show that the SimMIM method obtained the highest performance of 90.4\% and 88.9\% on detecting teeth and dental restorations and instance segmentation, respectively, increasing the average precision by 13.4 and 12.8 over the random initialization baseline. Moreover, we augment and correct the existing dataset of panoramic radiographs.
   The code and the dataset are available at https://github.com/AmaniHAlmalki/DentalMIM.
\end{abstract}

\section{Introduction}
The need for computer-assisted decisions is rising to facilitate diagnosis and treatment planning for dental care providers. Dental imaging is a valuable diagnostic tool for diagnosis and treatment plans, which is not possible solely through clinical exams and patient history \cite{white2001parameters}. A dental panoramic X-ray is a comprehensive tool that screens the teeth, surrounding alveolar bone and upper and lower jaws \cite{rushton1996use}.

Moreover, dental restoration is a biocompatible synthetic material used to restore missing tooth structures. The missing tooth structure can be restored with full and partial coverage depending on the extension and intensity of the missing structure to restore the tooth's coronal (top) part. Furthermore, root canal filling is a restorative procedure used to fill the space inside the tooth structure (root portion) with biocompatible restorative materials. Various dental restorative materials are available in the dental world; each has its indication, advantages, disadvantages, and clinician preferences. Most dental restorative materials appear radiopaque in the x-ray, and they can be identified by dental care providers \cite{abdalla2020artificial,molander1996panoramic}.

However, manual intervention for teeth numbering and identification of tooth restorations is time-consuming and may overlook significant data. Thus, the interest in computer vision and computer science for automated processes was aroused. Few studies have attempted to apply computer vision algorithms in dental radiograph analysis.
They include convolutional neural networks (CNNs) for teeth numbering and instance segmentation \cite{litjens2017survey}, two-stage network \cite{zhao2020tsasnet}, Faster R-CNN \cite{chen2019deep,kim2020automatic,tuzoff2019tooth,simonyan2014very,zhang2018effective}, PANet \cite{silva2020study}, Mask R-CNN \cite{jader2018deep,lee2020application,leite2021artificial,chen2017deeplab}, and U-Net network \cite{ronneberger2015u,ronneberger2015dental,koch2019accurate}. Recently, CNNs have enormous emerging applications in analyzing medical images with the advent of computation hardware/algorithm and expansion in the amount of data \cite{litjens2017survey}. However, CNNs are limited in overall capability because of inherent inductive biases \cite{dosovitskiy2020image}.

In this study, we propose to use a recently introduced Swin Transformer \cite{liu2021swin}
to analyze dental panoramic radiographs.
However, Swin Transformer requires large data for training, but there is only a very limited number of available dental radiographs. To alleviate this problem, we propose to use self-supervised learning.  To the best of our knowledge, this is the first study that applied self-supervised learning methods to Swin Transformer on dental panoramic radiographs.

Recently, the self-supervised learning methods, SimMIM \cite{xie2022simmim}, UM-MAE \cite{li2022uniform}, BEiT \cite{bao2021beit}, MAE \cite{he2022masked}, SplitMask \cite{el2021large}, MoCo v3 \cite{chen2021empirical}, and DINO \cite{caron2021emerging}, are effective in pre-training Transformers \cite{dosovitskiy2020image,liu2021swin} for learning visual representation. However, only UM-MAE and SimMIM pre-training methods are enabled for Pyramid-based ViTs with locality (Swin Transformer). Generally, the Masked Image Modeling (MIM) methods mask some image patches before they are fed into the transformer to predict the original patches in the masked area. This feature of aggregating information from the context helps many vision tasks. Although both UM-MAE and SimMIM provide a simple and efficient pre-training strategy for the Swin transformer encoder \cite{liu2021swin}, the process of the input to the encoder is dissimilar. MAE discards the masked tokens and inputs only visible patches to the lightweight decoder. However, MAE also breaks the two-dimensional structure of the input image. Therefore, it is not applicable to the Swin transformer without the Uniform Masking (UM) introduced in \cite{li2022uniform} to bridge the gap between the MAE and Swin transformer. SimMIM includes the masked tokens in the encoder and uses them as a direct prediction mechanism. Using the randomly masked patches for SimMIM is a reasonable reconstruction target, and a lightweight prediction head is sufficient for pre-training.
In addition, the location of the patches is essential in dental radiographs for a predictable outcome. SimMIM maintains the location of the patches known to both encoder and decoder, while MAE drops the location information, which may induce inaccuracy, as we demonstrate in this paper. 

As there is no standard dental image dataset for pre-training (unlike ImageNet for natural images), SimMIM and UM-MAE are trained on the same dataset as the downstream tasks (excluding the test dataset). We conduct experiments on dental image tasks, including teeth numbering, detection of dental restorations, and instance segmentation on the dental panoramic X-rays dataset \cite{silva2020study}.
For these tasks, we use the base Swin Transformer (Swin-B) \cite{liu2021swin} as the backbone of Cascade Mask R-CNN \cite{cai2019cascade}. We compare four Swin Transformer initializations, including SimMIM and UM-MAE, supervised initialization, and random initialization baseline. Our results show that SimMIM self pre-training can significantly improve object detection and instance segmentation performance on dental images. 

Although previous studies have investigated teeth segmentation, we still address many gaps in this work. First, there is no comprehensive instance segmentation data set for teeth numbering. Previous work on the matter \cite{silva2020study} used modified versions of binary semantic segmentation masks, which leads to a lack of instance overlapping and low-resolution outputs, resulting in inaccurate predictions, especially on the boundaries of the teeth. Second, there is a considerable amount of systematic errors because of the absence of dental expert supervision. Third, no prior work has simultaneously considered dental restoration segmentation besides tooth segmentation. The inclusion of teeth restorations increases the complexity of the computer vision problem because of class quantity and class imbalance.

To solve the data set issues, we augment and correct the existing dataset introduced in \cite{silva2020study}. 
In addition to correcting the manual segmentation errors under expert supervision, we further expand the dataset by developing annotations for dental restorations, including direct restorations, indirect restorations, and root canal therapy. The labeling procedure resulted in a unique high quality, augmented dataset. Our data is available, upon request, under the name TNDRS (Teeth Numbering, Detection of Restorations, and Segmentation) annotations.

Our main contributions are twofold: 
\begin{itemize}
\item We utilize self-supervised learning with SimMIM and UM-MAE to alleviate the problem of small data for panoramic radiographs.
\item The corrected dataset leads to a significant increase in performance, while added labeling of dental restorations extends the horizon of possible dental applications.
\end{itemize}
\section{Teeth numbering}
In dentistry, various dental numbering systems are available for teeth numbering for adults and children. These numbering systems are universally accepted for better communication between dental care providers. The Universal Numbering System, Palmer Notation Numbering System, and Federation Dentiaure International numbering system (FDI) are the most commonly used system across the globe among dental professionals. The FDI system is the most widely used international system. In this system, every single tooth is assigned two-digit numbers; the first digit number represents each quadrant. The maxillary right and left quadrants are identified by the numbers 1 and 2, while the mandibular left and right quadrants are the numbers 3 and 4, respectively. The second digit numbers represent each tooth based on its location in the jaw from the middle. The central incisor is assigned to number 1, whereas the third molar is set to number 8 \cite{tuzoff2019tooth,smith1976numbering}.

\graphicspath{{images/}}

\section{Methods}
\label{sec:init}

\begin{figure*}
\begin{center}
\includegraphics[width=1\linewidth]{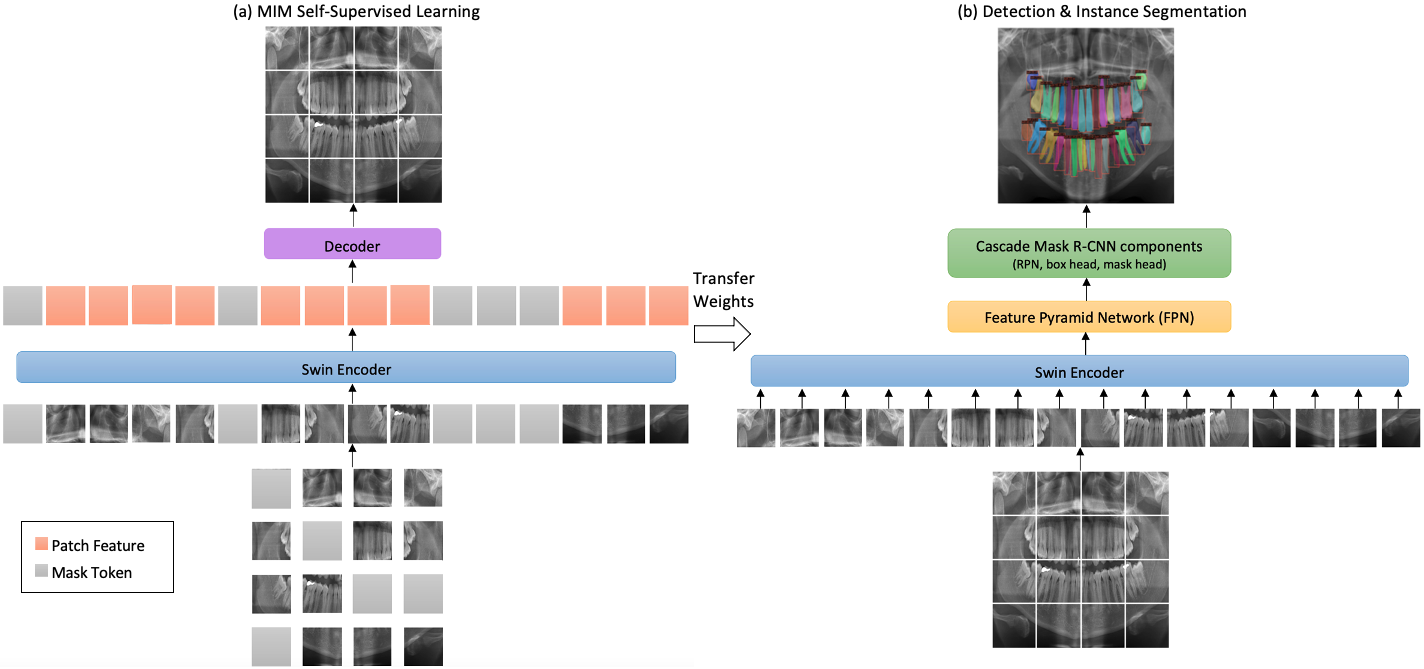} 
\end{center}
   \caption{Pipeline for teeth detection, detection of dental restorations, and instance segmentation with MIM Self Pre-training. (a) A Swin Transformer is first pre-trained by MIM methods on the target dataset. (b) The pre-trained Swin Transformer is used as the backbone in Cascade Mask R-CNN with FPN for the detection and segmentation tasks.}
\label{fig:pipeline}
\end{figure*}

The methods include two stages: the MIM pre-training and the downstream tasks, as illustrated in Fig.~\ref{fig:pipeline}.

In the first stage, Swin Transformer is pre-trained with MIM self-supervised learning methods as the encoder.
SimMIM divides the image into patches, replacing some random patches with mask tokens. Then, these patches, along with mask tokens, are input to the Swin encoder.
Hence the positional encoding of both visible and masked patches is preserved,
while UM-MAE drops those mask positions entirely. UM-MAE samples three random patches from each two-by-two grid, dropping 25\% of the entire image. Then it randomly masks 25\% of the already sampled areas as shared learnable tokens. Finally, the sampled patches and the masked tokens are reorganized as a compact two-dimensional input under a quarter of the original image resolution to feed via the Swin encoder.

\begin{figure*}
\begin{center}
\includegraphics[width=.9\linewidth]{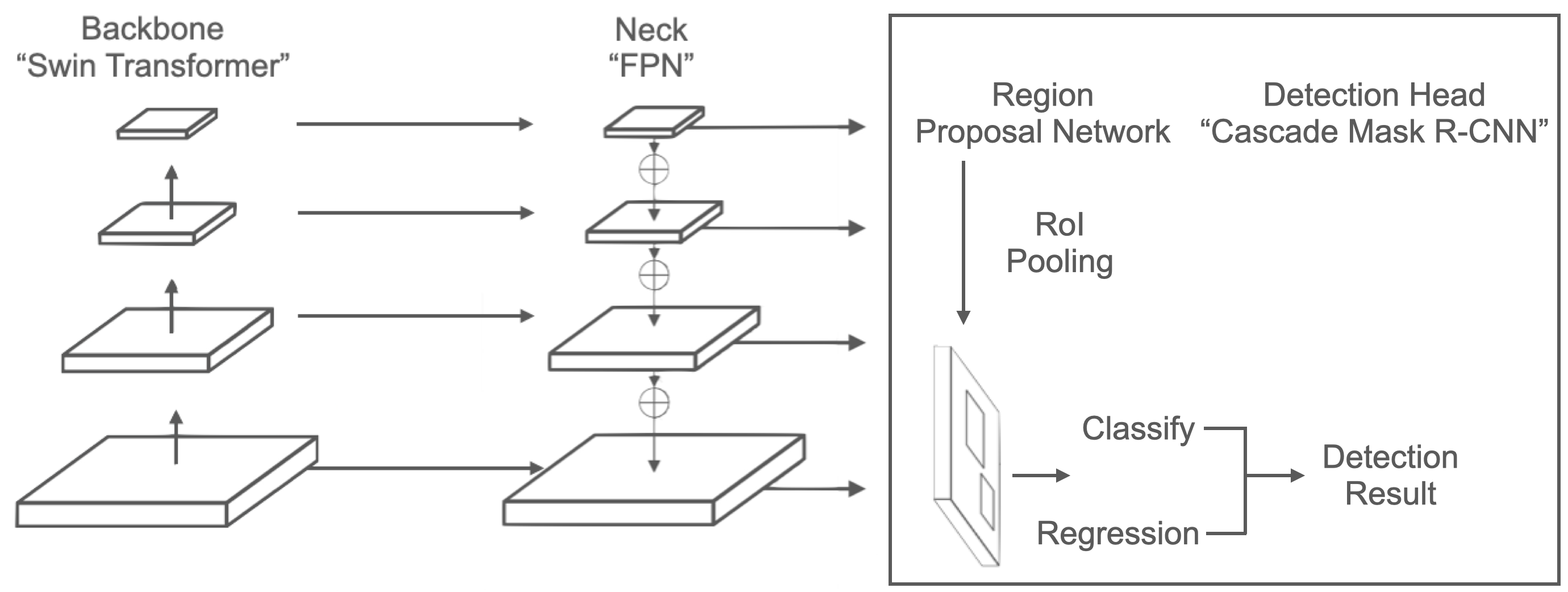} 
\end{center}
   \caption{Illustration of the architecture for object detection.}
\label{fig:cascade}
\end{figure*}

Then a decoder is appended to reconstruct the original patches at the masked area for both methods. In the second stage, the pre-trained Swin weights are transferred to initialize the detection and segmentation encoder. The features of the Swin Transformer backbone are fed to the neck (FPN \cite{lin2017feature}) and detection head (Cascade Mask R-CNN) for bounding box regression and classification as illustrated in Fig.~\ref{fig:cascade}. We select the Cascade Mask R-CNN \cite{cai2019cascade} framework due to its ubiquitous presence in object detection and instance segmentation research. Then, the whole network is fine-tuned to perform the detection and segmentation tasks.

We use the base Swin Transformer backbone (Swin-B) and compare the effectiveness of four configurations as follows:

{\bf Random.} The network is trained from scratch with randomly initialized weights, and no self-supervised methods are used. The Swin backbone configuration follows the code of \cite{liu2021swin}, and the Cascade Mask R-CNN configuration uses the defaults in MMDetection \cite{chen2019mmdetection}.

{\bf Supervised.} The Swin backbone is pre-trained for supervised object detection and instance segmentation using ImageNet-1K \cite{deng2009imagenet} images with their labels. We use the weights from \cite{liu2021swin} for Swin-B. Swin-B was pre-trained for 300 epochs. 

{\bf SimMIM.} We use the Swin-B weights pre-trained on self-supervised ImageNet-1K from \cite{xie2022simmim}. This model was pre-trained for 100 epochs.

{\bf UM-MAE.} Since ImageNet-1K pre-trained weights are not available; we use the official UM-MAE code release \cite{li2022uniform} to train Swin-B ourselves for 800 epochs (the default training length used in \cite{li2022uniform}) on unsupervised ImageNet-1K.

\section{Experiments}

\subsection{Dataset augmentation and correction}
\label{sec:data}

\begin{figure*}
\begin{center}
\includegraphics[width=.8\linewidth]{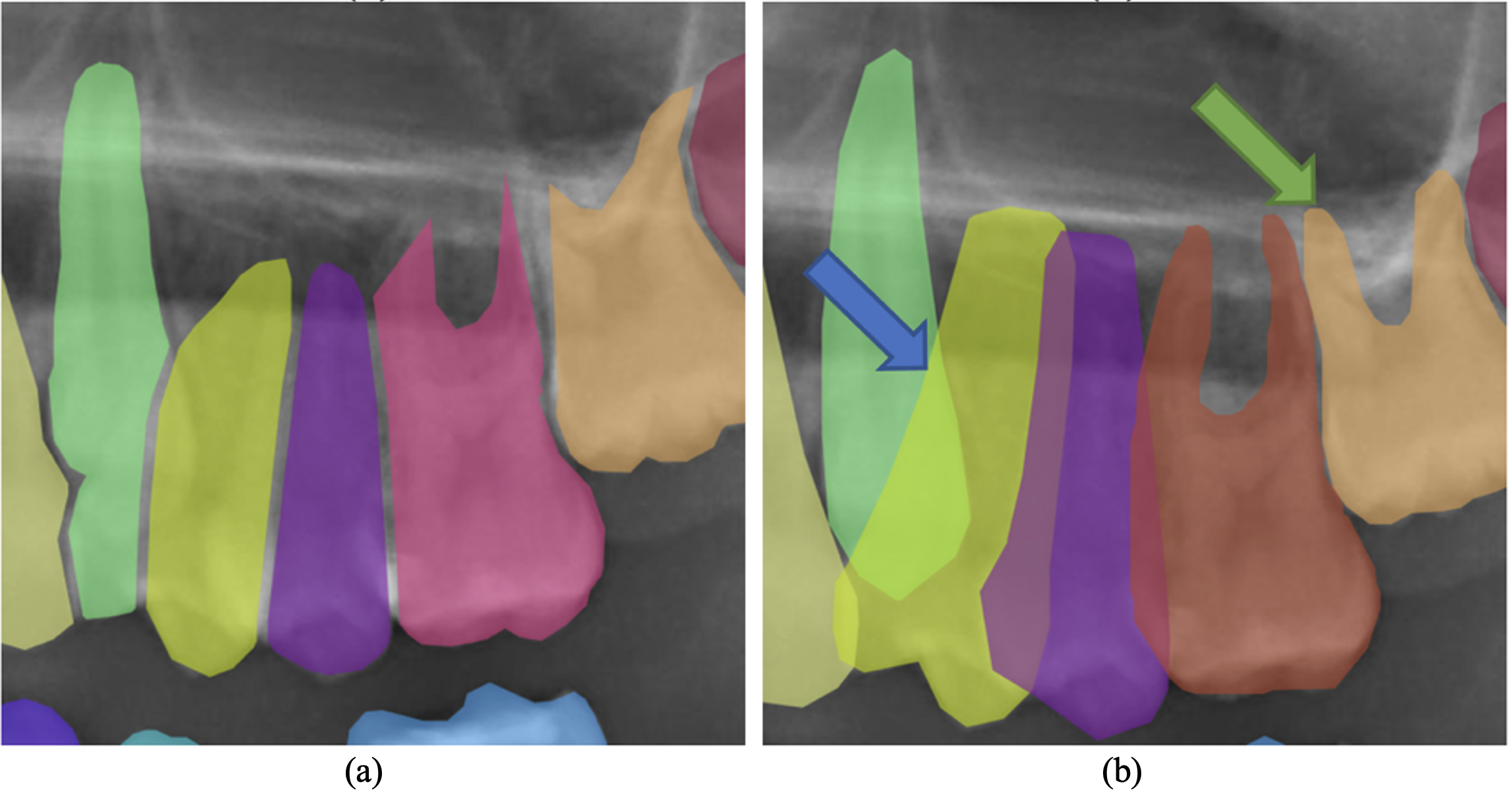}
\end{center}
   \caption{Comparison between the old and new dataset annotations. (a) Dataset old annotations. (b) Dataset new annotations. The blue arrow donates the inclusion of instance overlapping, while the green arrow indicates the correction of systematic errors, for example, unsegmented molar roots.}
\label{fig:modification}
\end{figure*}

\begin{figure*}
\begin{center}
\includegraphics[width=1\linewidth]{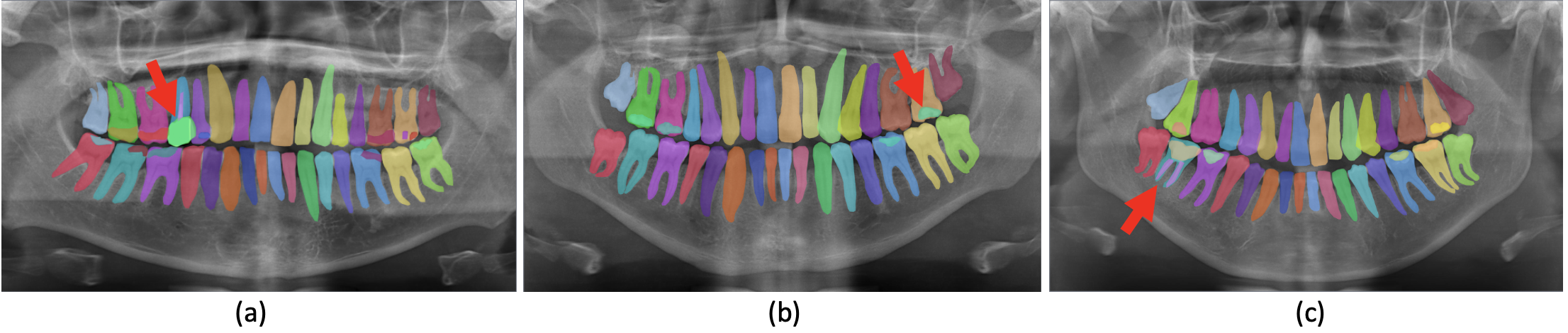} 
\end{center}
   \caption{Samples of segmentation of dental restorations. Red arrows show an example of a) indirect restoration, b) direct restoration, and c) root canal therapy.}
\label{fig:exampleRes}
\end{figure*}

{\bf TNDRS dental panoramic radiographs dataset.} Detection, Numbering, and Segmentation (DNS) \cite{silva2020study} is a dental panoramic X-rays dataset consisting of 543 annotated images with ground truth segmentation labels, including numbering information based on the FDI teeth numbering system. The image size is 1991x1127 pixels. The dataset annotations have some limitations as follows: 1) lack of instance overlapping; 2) some systematic errors because of the absence of dental expert supervision; 3) no segmentation of dental restorations. To overcome these issues, we modify and correct teeth instance segmentation and overlapping in all images. In addition, we contribute to further expanding the dataset by developing segmentation for dental restorations, including direct restorations, indirect restorations, and root canal therapy. This process was under a supervision of a dentist using the COCO-Annotator tool \cite{cocoannotator}. We attended weekly meetings where related issues, such as numbering, dental restorations, and segmentation questions, were discussed. In the end, the annotations were reviewed to assure quality and avoid systematic and random errors. Fig.~\ref{fig:modification} shows a sample comparing the old and new versions of the dataset annotations, highlighting both the instance overlapping (blue arrow) and the correction of systematic errors (green arrow). Fig.~\ref{fig:exampleRes} presents samples of segmentation of dental restorations. 

We believe this is the most inclusive dataset for segmenting teeth and dental restorations in dental panoramic radiographs. We are providing our data, upon request, under the name TNDRS (Teeth Numbering, Detection of Restorations, and Segmentation) annotations.

\subsection{Evaluation metric}
For all our experiments, we split the data into five folds, each containing approximately 20\% of the images. One of these folds is fixed as the test dataset (consisting of 111 images), and the other four folds (consisting of 108 images each) compose the training and validation datasets in a cross-validation manner. This process is repeated five times. The evaluation metric we adopt is the Average Precision for object detection and instance segmentation models.

\subsection{Implementation details}
Our experiments are implemented based on the PyTorch \cite{paszke2019pytorch} framework and trained with NVIDIA Tesla Volta V100 GPUs. In all experiments, the batch size equals the total number of the training sample, which is 432. The input images are all resized to 800×600 pixels. We utilize the AdamW \cite{loshchilov2017decoupled} optimizer in all experiments.

{\bf Data augmentation.} We apply noise addition and horizontal flipping, which changes teeth numbers to their equivalent new values (left teeth numbers turned into the right numbers and vice-versa).

{\bf SimMIM pre-training.} The base learning rate is set to 8e-4, weight decay is 0.05, $\beta1$ = 0.9, $\beta2$ = 0.999, with a cosine learning rate scheduler with warm-up for 10 epochs. We use a random MIM with a patch size of 16×16 and a mask ratio of 20\%. We employ a linear prediction head with a target image size of 800×600 and use L1 loss to compute the loss for masked pixel prediction.

{\bf UM-MAE pre-training.} The base learning rate is set to 1.5e-4, weight decay is 0.05, $\beta1$ = 0.9, $\beta2$ = 0.95, with a cosine decay learning rate scheduler with warm-up for 10 epochs. We use a random MIM with a patch size of 16×16 and a mask ratio of 25\%. We employ a linear prediction head with a target image size of 800×600 and adopt mean squared error (MSE) to compute the loss for masked pixel prediction.

{\bf Task fine-tuning.} For downstream tasks, we utilize single-scale training. The initial learning rate is 0.0001, and the weight decay is 0.05.

\section{Results and analysis}
\begin{figure*}
\begin{center}
\includegraphics[width=.8\linewidth]{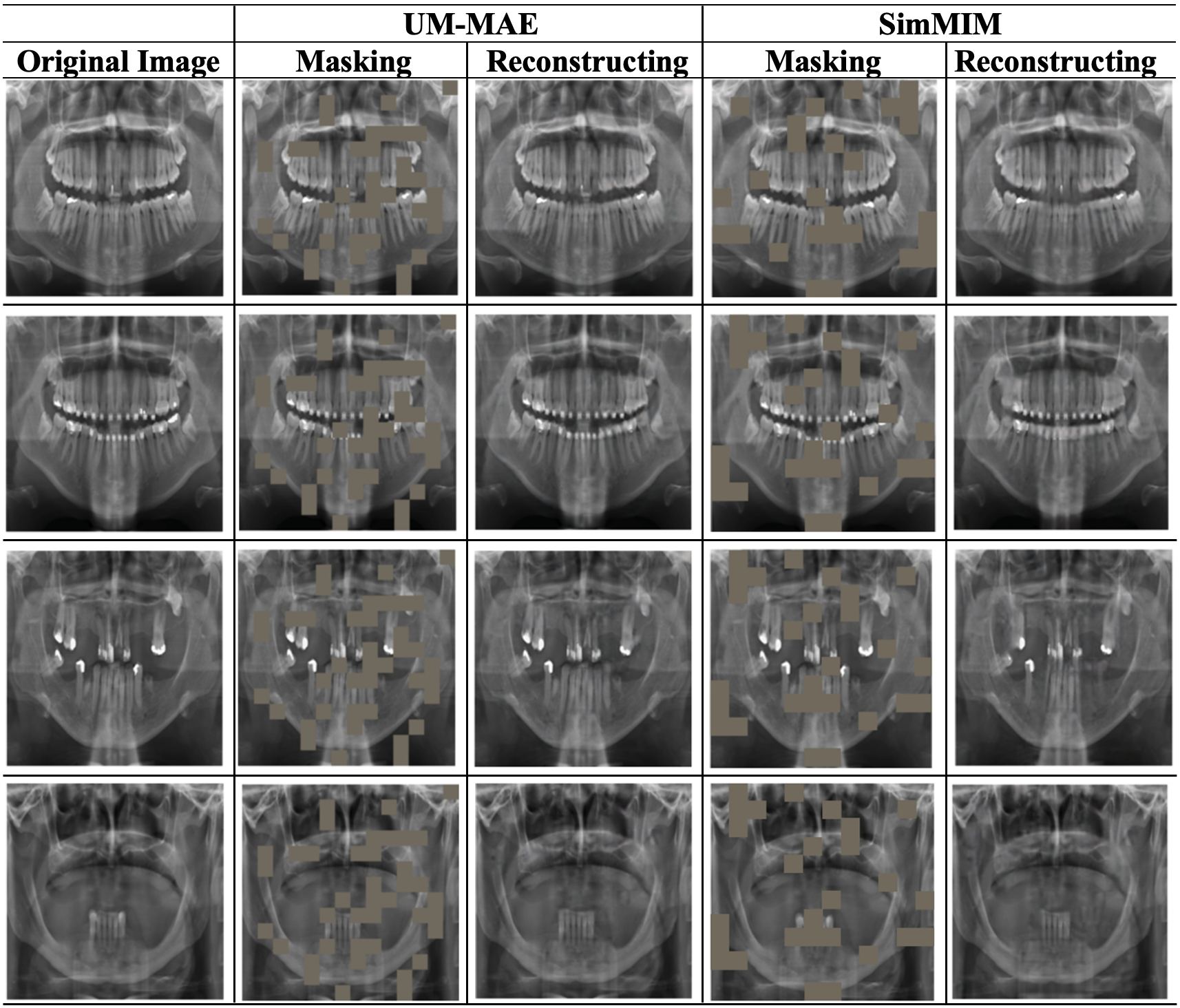} 
\end{center}
   \caption{SimMIM and UM-MAE reconstruction results. The first column is the original image, and the second and fourth columns are the masked image where the masked region is denoted by gray patches. The third and fifth columns are the reconstruction of MIM from the unmasked patches.}
\label{fig:reconst}
\end{figure*}

{\bf SimMIM and UM-MAE reconstruction.} The reconstruction results of SimMIM and UM-MAE are shown in Fig.~\ref{fig:reconst}. The five columns show the original images, the UM-MAE masked images, the UM-MAE reconstructed images, the SimMIM masked images, and the SimMIM reconstructed images. The results show that both MIM methods can restore lost information from the random context. It is worth noting that the ultimate goal of the MIM is to benefit the downstream tasks instead of generating high-quality reconstructions.

\subsection{Quantitative results}
\label{sec:quan}

{\bf Comparing initializations.} Table~\ref{table:results} shows the results of teeth detection and instance segmentation only and compares them to the previously published article from Silva \etal~\cite{silva2020study}. We present TNDRS fine-tuning results using the pre-trained models and random configurations described in Section~\ref{sec:init}. We make several observations.

(1) All four Swin Transformer initializations surpass the CNN-based SOTA of PANet with ResNet-50 backbone using ImageNet pre-training from Silva \etal~\cite{silva2020study}.

(2) Fine-tuning from supervised IN-1K pre-training yields 3.4 higher ${AP}^{box}$ than training from scratch (79.1 vs. 75.7) and 3.5 higher ${AP}^{mask}$ (78.3 vs. 74.8).

(3) UM-MAE substantially outperforms supervised initialization by 5.4 ${AP}^{box}$ (84.5 vs. 79.1), and 4.9 ${AP}^{mask}$ (83.2 vs. 78.3). 

(4) SimMIM outperforms UM-MAE by 1.6 ${AP}^{box}$ (86.1 vs. 84.5), and 1.4 ${AP}^{mask}$ (84.6 vs. 83.2). 

\begin{table*}
  \begin{center}
    {\small{
\begin{tabular}{lllll}
\toprule
Initialization & Backbone  & Pre-training Data & ${AP}^{box}$ & ${AP}^{mask}$ \\
\midrule
PANet\cite{silva2020study} & ResNet-50 & IN-1K w/ Labels   & 75.4  & 73.9   \\ \midrule
Random               & Swin-B    & None              & 75.7  & 74.8   \\
Supervised           & Swin-B    & IN-1K w/ Labels   & 79.1  & 78.3   \\
UM-MAE                  & Swin-B    & IN-1K             & 84.5  & 83.2   \\
SimMIM               & Swin-B    & IN-1K             & {\bf 86.1}  & {\bf 84.6}   \\ 
\bottomrule
\end{tabular}
}}
\end{center}
\caption{Results of teeth detection and instance segmentation only.}
\label{table:results}
\end{table*}

Table~\ref{table:resultsRes} compares the four Swin Transformer initializations after data augmentation of dental restorations. Our results prove that the SimMIM method achieved the highest performance of 90.4\% and 88.9\% on detecting teeth and dental restorations and instance segmentation, respectively.

\begin{table*}
  \begin{center}
    {\small{
\begin{tabular}{lllll}
\toprule
Initialization & Backbone  & Pre-training Data & ${AP}^{box}$ & ${AP}^{mask}$ \\
\midrule
Random               & Swin-B    & None              & 77.0  & 76.1   \\
Supervised           & Swin-B    & IN-1K w/ Labels   & 80.3  & 79.2   \\
UM-MAE                  & Swin-B    & IN-1K             & 88.3  & 85.7   \\
SimMIM               & Swin-B    & IN-1K             & {\bf 90.4}  & {\bf 88.9}   \\ 
\bottomrule
\end{tabular}
}}
\end{center}
\caption{Results after augmenting dental restorations.}
\label{table:resultsRes}
\end{table*}

{\bf Parameter setting.} In Table~\ref{table:maskings}, we conduct experiments on teeth detection and instance segmentation tasks with different SimMIM pre-training epochs and mask ratios. First, the performance of SimMIM does not benefit from longer training. Second, unlike the high mask ratio \cite{xie2022simmim} adopted in natural images, the downstream tasks show different preferences for the mask ratio. Both tasks are consistently improved with a decrease in mask ratio from 60\% to 10\%. The reason why this decrease facilitates the training may be attributed to the fact that the relevant features are small on panoramic X-rays.

\begin{table}
  \begin{center}
    {\small{
\begin{tabular}{llll}
\toprule
Mask ratio & Pre-training Epochs & ${AP}^{box}$ & ${AP}^{mask}$ \\
\midrule
60\% & 100 & 84.3   & 83.2  \\ \midrule
50\%               & 100    & 84.7              & 83.6   \\
50\%               & 800    & 83.1              & 83.0   \\ \midrule
40\%           & 100    & 85.5   & 83.9   \\
30\%                  & 100    & 85.9             & 84.1   \\
20\%               & 100    &  {\bf 86.1}  & {\bf 84.6}   \\ 
10\%                  & 100    & 85.8  & 84.3  \\
\bottomrule
\end{tabular}
}}
\end{center}
\caption{The influence of Mask Ratios on teeth detection and instance segmentation tasks.}
\label{table:maskings}
\end{table}

{\bf Dataset correction.} After we correct teeth segmentation on DNS discussed in Section~\ref{sec:data}, teeth detection and instance segmentation performance are remarkably improved by 5.9 ${AP}^{box}$ and 6.4 ${AP}^{mask}$ as shown in Table~\ref{table:improve}.

\begin{table}
  \begin{center}
    {\small{
\begin{tabular}{lll}
\toprule
DNS Annotations & ${AP}^{box}$ & ${AP}^{mask}$ \\
\midrule
Before Correction & 80.2 & 78.2  \\ 
After Correction & {\bf 86.1}  & {\bf 84.6}  \\
\bottomrule
\end{tabular}
}}
\end{center}
\caption{Correction of teeth segmentation.}
\label{table:improve}
\end{table}

\subsection{Qualitative results} 
\begin{figure*}
\begin{center}
\includegraphics[width=1\linewidth]{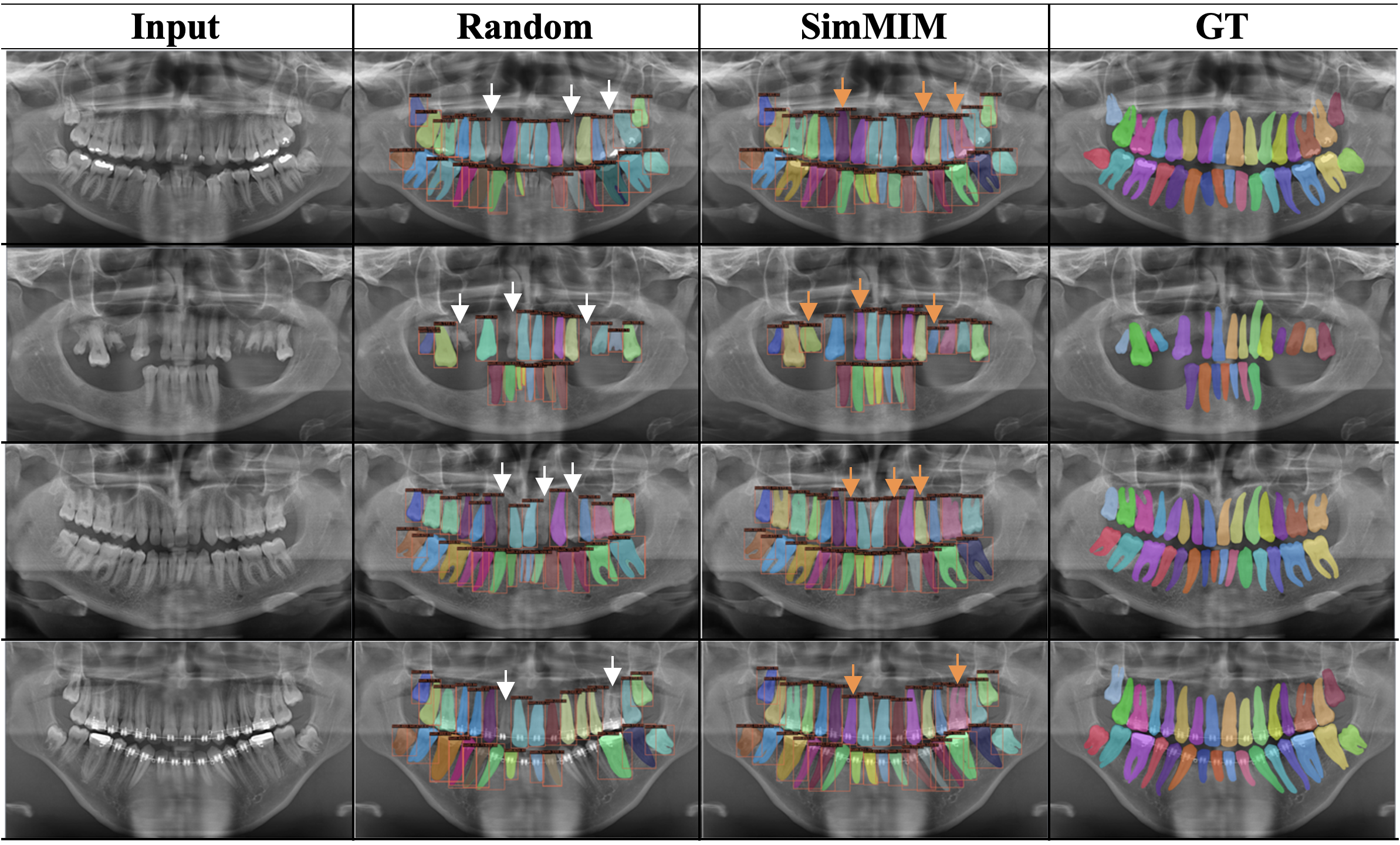} 
\end{center}
   \caption{Qualitative results of teeth detection and instance segmentation only. Note that teeth detection and instance segmentation are missing (white arrows) when created by the baseline Swin Transformer approach compared to the segmentation produced by Swin Transformer pre-trained with SimMIM architecture (orange arrows).}
\label{fig:qual}
\end{figure*}

In Fig.~\ref{fig:qual}, the displayed results for four different images demonstrate qualitative samples of improved performance when Swin Transformer is pre-trained with SimMIM for teeth detection and segmentation only. These improvements in detection and segmentation agree with the quantitative results in Section~\ref{sec:quan}.

\begin{figure*}
\begin{center}
\includegraphics[width=.7\linewidth]{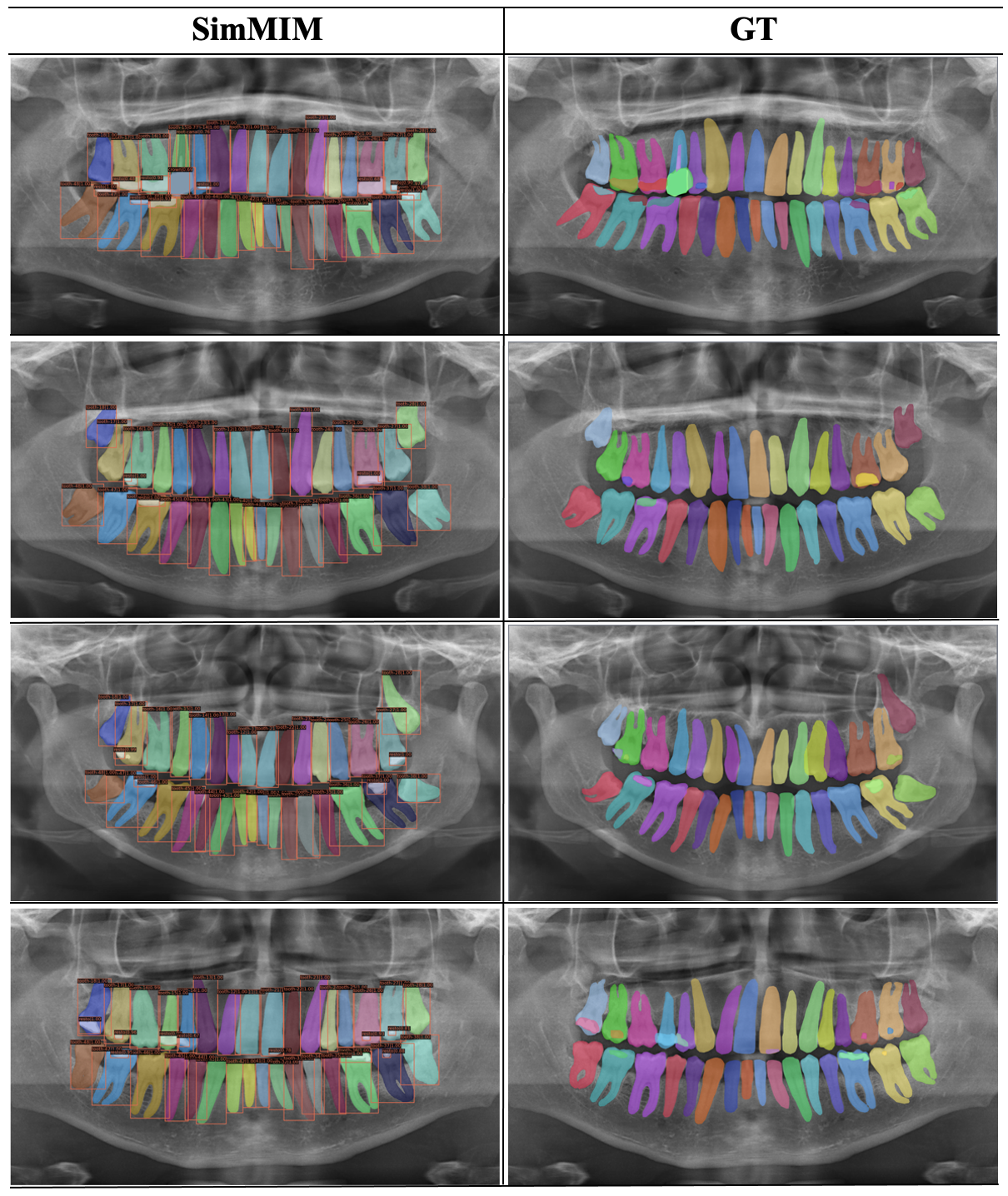} 
\end{center}
   \caption{Qualitative results of detecting teeth and dental restorations and instance segmentation using SimMIM.}
\label{fig:qualreS}
\end{figure*}

Fig.~\ref{fig:qualreS} displays qualitative results after augmenting dental restorations when Swin Transformer is pre-trained with SimMIM.

\subsection{Pre-training time and memory consumption}
Comparing UM-MAE to the SimMIM framework, the core advantage of UM-MAE is the memory and runtime efficiency. In Table~\ref{table:time}, we show their clear comparisons based on Swin-B. It is observed that UM-MAE speeds up by about 2× and reduces the memory by at least 2× against SimMIM, where their performances under the downstream tasks show the opposite.

\begin{table}
  \begin{center}
    {\small{
\begin{tabular}{lll}
\toprule
Method & Time & Memory \\
\midrule
SimMIM & 24.6 h & 18.7 GB \\ 
UM-MAE & {\bf 12.5 h} & {\bf 6.7 GB} \\

\bottomrule
\end{tabular}
}}
\end{center}
\caption{The comparison of pre-training time and memory consumption.}
\label{table:time}
\end{table}

\section{Conclusions}
Two self-supervised learning methods were applied to Swin Transformer on dental panoramic radiographs: SimMIM and UM-MAE. The results of the masking-based method, SimMIM, obtained superior performance than UM-MAE, supervised and random initialization for detection of teeth, dental restorations, and instance segmentation. 
Based on this experiment, we can conclude that adjusting parameters, including mask ratio and pre-training epochs, is useful when applying SimMIM pre-training to the dental imaging domain for reliable outcomes. 
In addition, correcting the dataset annotations lead to further improvements that significantly surpass the available state-of-the-art results. Our plan for future work is to examine the efficacy of SimMIM pre-training in prognosis and outcome prediction tasks.

\section{Acknowledgments}
We would like to express our deepest thanks to Dr. \mbox{Abdulrahman} Almalki, a  dental expert from the University of Pennsylvania, for his valuable discussions related to dentistry.
This work was in part supported by the NSF Grant IIS-1814745.  The calculations were carried out on HPC resources at Temple University supported in part by the NSF
through grant number 1625061 and by the U.S. ARL under
contract number W911NF-16-2-0189.

{\small
\bibliographystyle{ieee_fullname}
\bibliography{egbib}
}

\end{document}